\documentclass[journal]{IEEEtran}
\usepackage{amssymb}
\usepackage{pifont,calc}
\usepackage{cite}

\usepackage{graphicx}
\ifCLASSINFOpdf
   \usepackage[pdftex]{graphicx}
\else
 \fi
\usepackage[cmex10]{amsmath}
\usepackage{array}

\usepackage{cases} % to combine equations together
\usepackage{bm} % to make xila zimu hei ti
\usepackage{subfigure}
\usepackage{mathrsfs}
\usepackage{url}
\usepackage{algorithm}
\usepackage{algorithmic}
\usepackage{multirow}
\makeatletter

\newcommand{\Rmnum}[1]{\expandafter\@slowromancap\romannumeral #1@}
\makeatother

\ifCLASSINFOpdf
\else
\fi
\begin{document}
\title{Efficient Inverse-Free Algorithms for Extreme Learning Machine Based on the Recursive Matrix Inverse and the Inverse  ${\bf{LD}}{{\bf{L}}^T}$  Factorization}

\author{Hufei~Zhu  and Chenghao~Wei
\thanks{H. Zhu and C. Wei are with the College of Computer Science and Software, Shenzhen University, Shenzhen 518060, China (e-mail:
zhuhufei@szu.edu.cn; chenghao.wei@szu.edu.cn).}% <-this % stops a space
}


\markboth{Journal of \LaTeX\ Class Files,~Vol.~14, No.~8, August~2015}%
{Shell \MakeLowercase{\textit{et al.}}: Bare Demo of IEEEtran.cls for IEEE Journals}

\maketitle

\begin{abstract}
The inverse-free extreme learning machine (ELM) algorithm proposed in \cite{Existing_Inverse_free_best_ELM}
%can obtain the output weights that are identical to the optimal solution of the standard ELM with Tikhonov regularization. It
was based on an inverse-free algorithm to compute the regularized pseudo-inverse,  which was deduced
from an inverse-free recursive algorithm to update the inverse of a Hermitian matrix.
Before that recursive algorithm was applied in \cite{Existing_Inverse_free_best_ELM},
its improved version had been utilized  in previous literatures~\cite{zhfICC2009, TransComm2010ReCursiveGstbc}.
Accordingly from the improved recursive algorithm \cite{zhfICC2009, TransComm2010ReCursiveGstbc},
 we deduce a more efficient  inverse-free algorithm to update the regularized pseudo-inverse, from which we develop the proposed inverse-free ELM algorithm $1$.
 Moreover, the proposed ELM algorithm $2$ further reduces the computational complexity,  which
  computes  the  output weights directly from the updated inverse, and avoids computing the regularized pseudo-inverse.
Lastly,  instead of updating the inverse,
the proposed ELM algorithm $3$ updates the ${\bf{LD}}{{\bf{L}}^T}$
   factor of the inverse by  the inverse ${\bf{LD}}{{\bf{L}}^T}$ factorization \cite{zhfVTC2010DivFree},
   to avoid numerical instabilities after a very large number of
iterations~\cite{TransSP2003Blast}.  With respect to the
existing ELM algorithm, the proposed ELM algorithms $1$, $2$ and $3$ are expected to require only
%$\frac{2}{4+M}$, $\frac{1}{4+M}$ and $\frac{1}{4+M}$
$\frac{3}{8+M}$, $\frac{1}{8+M}$ and $\frac{1}{8+M}$
of
complexities, respectively, where $M$ is the output node number.
In the numerical experiments, the standard ELM, the
existing inverse-free ELM algorithm and the proposed ELM
algorithms 1, 2 and 3 achieve the same performance in regression and
classification,  while all the $3$ proposed algorithms
significantly accelerate the existing
inverse-free ELM algorithm. 
%in \cite{Existing_Inverse_free_best_ELM}.
\end{abstract}


% \fi
\IEEEpeerreviewmaketitle

\section{Introduction}

The extreme learning machine (ELM)~\cite{ELM2004cof}
is an
effective solution
 for Single-hidden-layer feedforward networks (SLFNs)
 due to its unique characteristics,
i.e., extremely fast learning speed, good generalization performance, and
universal approximation capability~\cite{ref17ElmInvFreePaper}.
Thus ELM
has
been
widely applied in classification and
regression~\cite{ELM2012RegressionAndClassify}.

The incremental ELM proposed in \cite{ref17ElmInvFreePaper}
%which can
achieves the universal approximation capability by
adding hidden nodes one by one.
However, it only updates the output weight  for
the newly added hidden node, and freezes the output weights of the existing hidden nodes.
Accordingly  those output weights
%%obtained by the inverse-free algorithm in \cite{ref17ElmInvFreePaper} must be
%are different from those computed
%by the conventional ELM with the inverse
%operation, and then
 are no longer the optimal least-squares solution of the standard ELM algorithm.
 Then the inverse-free algorithm was proposed in \cite{Existing_Inverse_free_best_ELM}
to  update the output weights of  the added node and the existing nodes simultaneously,
 and the updated weights are identical to the optimal solution of the standard ELM algorithm.
 The ELM algorithm in \cite{Existing_Inverse_free_best_ELM} was based on
an inverse-free algorithm to compute the regularized pseudo-inverse,  which was deduced
from an inverse-free recursive algorithm to update the inverse of a Hermitian matrix.

%The inverse-free recursive algorithm in \cite{Existing_Inverse_free_best_ELM} to update the inverse
%has been ,
Before the recursive algorithm to update the inverse was utilized in \cite{Existing_Inverse_free_best_ELM},
it had been mentioned in previous literatures~\cite{Matrix_Handbook,Matrix_SP_book,GlobecommRecursiveVBLAST, TransWC2009ReCursiveBlast,zhfICC2009},
%introduced in \cite{Matrix_Handbook,Matrix_SP_book} and applied in \cite{GlobecommRecursiveVBLAST, TransWC2009ReCursiveBlast},
while its improved version had been utilized  in \cite{zhfICC2009, TransComm2010ReCursiveGstbc}.
%has been simplified in \cite{zhfICC2009}.
Accordingly from the improved recursive algorithm  \cite{zhfICC2009, TransComm2010ReCursiveGstbc},
 we
deduce a more efficient  inverse-free algorithm to update the regularized pseudo-inverse, from which we develop the proposed ELM algorithm $1$. Moreover, the proposed ELM algorithm $2$  computes  the  output weights directly from the updated inverse,
%by the recursive algorithm in \cite{zhfICC2009},
to further reduce the computational complexity by avoiding the calculation of the regularized pseudo-inverse.
Lastly,  instead of updating the inverse,
the proposed ELM algorithm $3$ updates the ${\bf{LD}}{{\bf{L}}^T}$
   factors of the inverse by  the inverse ${\bf{LD}}{{\bf{L}}^T}$ factorization proposed in \cite{zhfVTC2010DivFree},
   since the recursive algorithm to update the inverse
may introduce numerical instabilities in the processor units with the finite precision, which occurs only after a very large number of
iterations~\cite{TransSP2003Blast}.
%when the processor units are limited in precision.

This correspondence is organized as follows. Section \Rmnum{2} describes the
ELM model. Section \Rmnum{3} introduces the existing inverse-free ELM algorithm~\cite{Existing_Inverse_free_best_ELM}.
 In Section \Rmnum{4}, we
deduce the proposed $3$ inverse-free ELM algorithms, and compare the expected computational complexities of the existing and proposed algorithms.
Section \Rmnum{5} evaluates the existing and proposed algorithms by numerical experiments.
Finally, we make conclusion in Section \Rmnum{6}.

\section{Architecture of the ELM}

In the ELM model, the
 $n$-th input node, the $i$-th hidden node, and the $m$-th output node can be denoted as $x_n$, ${h_i}$, and ${z_m}$, respectively,
 while all the $N$ input nodes, $l$ hidden nodes, and $M$ output nodes
 can be denoted as ${\bf{x}} = {\left[ {\begin{array}{*{20}{c}}
{{x_{1}}}&{{x_{2}}}& \cdots &{{x_{N}}}
\end{array}} \right]^T} \in {\Re ^N}$,
 ${\bf{h}} = {\left[ {\begin{array}{*{20}{c}}
{{h_{1}}}&{{h_{2}}}& \cdots &{{h_l}}
\end{array}} \right]^T} \in {\Re ^l}$,
and ${\bf{z}} = {\left[ {\begin{array}{*{20}{c}}
{{z_{1}}}&{{z_{2}}}& \cdots &{{z_{M}}}
\end{array}} \right]^T} \in {\Re ^M}$, respectively.
Accordingly
 the ELM model  can be represented  in a compact form
as
 \begin{equation}\label{equ03}
{\bf{h}} = f\left( {{\bf{Ax}} + {\bf{d}}} \right)
 \end{equation}
 and
  \begin{equation}\label{equ04}
{\bf{z}} = {\bf{Wh}},
 \end{equation}
 where
${\bf{A}} = \left[ {{a_{in}}} \right] \in {\Re ^{l \times N}}$,
${\bf{d}} = {\left[ {\begin{array}{*{20}{c}}
{{d_1}}&{{d_2}}& \cdots &{{d_l}}
\end{array}} \right]^T} \in {\Re ^l}$,
${\bf{W}} = \left[ {{w_{mi}}} \right] \in {\Re ^{M \times l}}$,
and  the activation
function $f( \bullet )$ is
%defined in
entry-wise, i.e., $f({\bf{A}}) = \left[ {f({a_{in}})} \right] \in {\Re ^{l \times N}}$ for a matrix input ${\bf{A}} = \left[ {{a_{in}}} \right] \in {\Re ^{l \times N}}$.
In (\ref{equ03}),  the activation
function $f( \bullet )$
%$f\left( {\sum\limits_{n = 1}^N {{a_{in}}{x_n}}  + {b_i}} \right)$ is an activation
%function, which
can be chosen as linear, sigmoid, Gaussian
models, etc.

Assume there are totally $K$ distinct   training samples, and let ${{\bf{x}}_{k}}\in {\Re ^N}$ and ${{\bf{z}}_{k}}\in {\Re ^M}$ denote the $k$-th training input and the corresponding $k$-th training output, respectively, where $k=1,2,\cdots,K$. Then the input sequence and the output sequence in the training set can be represented as
\begin{equation}\label{equ05}
{\bf{X}} = \left[ {\begin{array}{*{20}{c}}
{{{\bf{x}}_{1}}}&{{{\bf{x}}_{2}}}& \cdots &{{{\bf{x}}_{K}}}
\end{array}} \right] \in {\Re ^{N \times K}},
\end{equation}
and
 \begin{equation}\label{equ06}
{\bf{Z}} = \left[ {\begin{array}{*{20}{c}}
{{{\bf{z}}_{1}}}&{{{\bf{z}}_{2}}}& \cdots &{{{\bf{z}}_{K}}}
\end{array}} \right] \in {\Re ^{M \times K}},
 \end{equation}
respectively. We can substitute (\ref{equ05}) into (\ref{equ03}) to obtain
 \begin{equation}\label{equ07}
{\bf{H}} = f\left( {{\bf{AX}} + {{\bf{1}}^T} \otimes {\bf{d}}} \right),
  \end{equation}
where ${\bf{H}} = \left[ {\begin{array}{*{20}{c}}
{{{\bf{h}}_{1}}}&{{{\bf{h}}_{2}}}& \cdots &{{{\bf{h}}_{K}}}
\end{array}} \right] \in {\Re ^{l \times K}}$ is the value sequence of all $l$ hidden nodes,
and $\otimes$ is the Kronecker product~\cite{Existing_Inverse_free_best_ELM}. Then we can substitute (\ref{equ07}) and (\ref{equ06})
%and (\ref{equ07})
into  (\ref{equ04}) to obtain the actual training
output sequence
\begin{equation}\label{equ08}
{\bf{Z}} = {\bf{WH}}.
\end{equation}

In an ELM, only the output weight ${\bf{W}}$ is adjustable, while ${\bf{A}}$ (i.e., the input weights) and ${\bf{d}}$ (i.e., the biases of the hidden nodes) are randomly fixed.
Denote the desired output as ${\bf{Y}}$.
Then an ELM
simply minimizes the estimation error
\begin{equation}\label{Edefine32178a}
{\bf{E}}={\bf{Y}} - {\bf{Z}}={{\bf{Y}} - {\bf{WH}}}
\end{equation}
by finding a least-squares solution ${\bf{W}}$ for the problem
 \begin{equation}\label{equ09}
\mathop {\min }\limits_{\bf{W}} \left\| {\bf{E}} \right\|_F^2=\mathop {\min }\limits_{\bf{W}} \left\| {{\bf{Y}} - {\bf{WH}}} \right\|_F^2,
 \end{equation}
where
$\left\| {\bullet} \right\|_F$  denotes the Frobenius norm.

For the problem (\ref{equ09}), the unique minimum norm least-squares solution is~\cite{ELM2004cof}
%${\bf{W}} = {{\bf{H}}^\dag }{\bf{Y}}$, i.e.,
 \begin{equation}\label{equ10}
{\bf{W}} = {\bf{Y}}{{\bf{H}}^T}{\left( {{\bf{H}}{{\bf{H}}^T}} \right)^{ - 1}}.
 \end{equation}
To avoid over-fitting, the popular Tikhonov regularization~\cite{48_regularized,49_regularized} can
be utilized to  modify (\ref{equ10}) into
%then (\ref{equ10}) should be modified into
 \begin{equation}\label{equ11}
{\bf{W}} = {\bf{Y}}{{\bf{H}}^T}{\left( {{\bf{H}}{{\bf{H}}^T}{+}k_0^2{\bf{I}}} \right)^{ - 1}},
 \end{equation}
 where $k_0^2 >0$ denotes the regularization factor.
 %It can easily be  seen that
 Obviously  (\ref{equ10}) is just the special case of (\ref{equ11}) with $k_0^2=0$. Thus in what follows,
 we only consider (\ref{equ11}) for the ELM with Tikhonov regularization.

\section{The Existing Inverse-Free ELM Algorithm}

In machine learning, it is a common strategy  to
 increase the hidden node number gradually until the desired
accuracy is achieved. However, when this strategy is applied in ELM directly, the matrix inverse operation in (\ref{equ11})
 %involved in
 for the conventional ELM will be required when a few or only one extra hidden node is introduced, and accordingly the algorithm will be computational prohibitive.
Accordingly  an inverse-free strategy was proposed in \cite{Existing_Inverse_free_best_ELM},
to  update the output weights incrementally with the increase of the hidden nodes. In each step,
 the output weights obtained by the inverse-free algorithm
 are identical to the solution of the standard ELM algorithm using the inverse operation.

 Assume that in the ELM with $l$ hidden nodes, we
 add
 %increase
 %only
 one extra hidden node, i.e., the hidden node $l+1$, which has the input weight row vector ${{\bf{\bar a}}_{l + 1}^T} = {\left[ {\begin{array}{*{20}{c}}
{{a_{(l + 1)1}}}&{{a_{(l + 1)2}}}& \cdots &{{a_{(l + 1)N}}}
\end{array}} \right]}$ $\in ({\Re ^N})^T$ and the bias ${\bar d}_{l+1}$.  Then from (\ref{equ07}) it can be seen that
the extra row ${\bf{\bar h}}_{l + 1}^T = f\left( {{\bf{\bar a}}_{l + 1}^T{\bf{X}} + {\bar d}_{l+1}{{\bf{1}}^T} } \right)$ needs to be added to
${\bf{H}} $, i.e.,
 \begin{equation}\label{Haddrow2134}
{{\bf{H}}^{l + 1}} = \left[ \begin{array}{l}
{{\bf{H}}^l}\\
{\bf{\bar h}}_{l + 1}^T
\end{array} \right],
 \end{equation}
 where  ${{\bf{H}}^{i}}$ ($i=l,l+1$)  denotes ${\bf{H}}$ for the ELM with $i$ hidden nodes.
In  ${{\bf{\bar a}}_{l + 1}}$, ${\bf{\bar h}}_{l + 1}$, ${\bar d}_{l+1}$ and what follows, we add the overline
%${\bar  \cdot }$
to emphasize the extra vector or scalar, which is added to the matrix or vector for the ELM with $l$ hidden nodes.

After ${\bf{H}} $ is updated by (\ref{Haddrow2134}), the conventional ELM updates the output weights by
%(\ref{equ10}) or
  (\ref{equ11}) that involves an inverse
operation.
 To avoid that inverse
operation, the algorithm in \cite{Existing_Inverse_free_best_ELM}
utilizes an inverse-free algorithm to update
 \begin{equation}\label{equ0defB3789}
{\bf{B}} = {{\bf{H}}^T}{\left( {{\bf{H}}{{\bf{H}}^T}{+}k_0^2{\bf{I}}} \right)^{ - 1}}
 \end{equation}
 that is the regularized pseudo-inverse of
 %the value
%sequence in all hidden nodes
${{\bf{H}}}$,
and then substitutes (\ref{equ0defB3789}) into (\ref{equ11}) to
 %to
 compute the output weights by
 \begin{equation}\label{equ0DefWbyB32797}
{\bf{W}} = {\bf{Y}}{\bf{B}}.
 \end{equation}
 In \cite{Existing_Inverse_free_best_ELM},
% and then compute
%and it computes
${{\bf{B}}^{l + 1}}$ (i.e.,
%the regularized pseudo-inverse
 ${\bf{B}}$ for the ELM with $l+1$ hidden nodes) is computed from ${{\bf{B}}^l}$ iteratively
 by
 %and the corresponding equations are
  \begin{equation}\label{equ0Bextend94375}
{{\bf{B}}^{l + 1}} = \left[ {\begin{array}{*{20}{c}}
{{{{\bf{\tilde B}}}^l}}&{{{\bf{\bar b}}_{l+1}}}
\end{array}} \right],
 \end{equation}
where
 \begin{multline}\label{equ0Bwave94375}
{{\bf{\tilde B}}^l} = \frac{{\left( {({{\bf{\bar h}}_{l+1}^T}{\bf{\bar h}}_{l+1}{+}k_0^2){\bf{I}} - {\bf{\bar h}}_{l+1}{{\bf{\bar h}}_{l+1}^T}} \right){{\bf{B}}^l}{{ {\bf{H}}^l {\bf{\bar h}}_{l+1}}}{{\bf{\bar h}}_{l+1}^T}{{\bf{B}}^l}}}{{({{\bf{\bar h}}_{l+1}^T}{\bf{\bar h}}_{l+1}{+}k_0^2)({{\bf{\bar h}}_{l+1}^T}{\bf{\bar h}}_{l+1}{+}k_0^2 - {{\bf{\bar h}}_{l+1}^T}{{\bf{B}}^l}{{\bf{H}}^l {\bf{\bar h}}_{l+1} })}} \\
+ \frac{{\left( {({{\bf{\bar h}}_{l+1}^T}{\bf{\bar h}}_{l+1}{+}k_0^2){\bf{I}} - {\bf{\bar h}}_{l+1}{{\bf{\bar h}}_{l+1}^T}} \right){{\bf{B}}^l}}}{{{{\bf{\bar h}}_{l+1}^T}{\bf{\bar h}}_{l+1}{+}k_0^2}},
 \end{multline}
and ${{\bf{\bar b}}_{l+1} }$,  the $(l+1)^{th}$ column of ${{\bf{B}}^{l + 1}}$,  is
computed by
 \begin{equation}\label{equ0B_bar94375}
{{\bf{\bar b}}_{l+1} }{{ = }} - \frac{{{{{\bf{\tilde B}}}^l} {\bf{H}}^l {\bf{\bar h}}_{l+1}}}{{{{\bf{\bar h}}_{l+1}^T}{\bf{\bar h}}_{l+1}{+}k_0^2}} + \frac{{\bf{\bar h}}_{l+1}}{{{{\bf{\bar h}}_{l+1}^T}{\bf{\bar h}}_{l+1}{+}k_0^2}}.
 \end{equation}

Let
 \begin{equation}\label{equ0Rdef43415}
{\bf{R}} = {\bf{H}}{{\bf{H}}^T}{+}k_0^2{\bf{I}}
 \end{equation}
and
 \begin{equation}\label{equ0QRrelation3123}
{\bf{Q}} =  {{\bf{R}}^{ - 1}}= {\left( {{\bf{H}}{{\bf{H}}^T}{+}k_0^2{\bf{I}}} \right)^{ - 1}}.
 \end{equation}
% \begin{equation}\label{equ0Qdefine31942}
%{\bf{Q}} = {\left( {{\bf{H}}{{\bf{H}}^T}{+}k_0^2{\bf{I}}} \right)^{ - 1}}
% \end{equation}
Then we can write (\ref{equ0defB3789}) as
 \begin{equation}\label{equ0defB3789ToQ}
{\bf{B}} = {{\bf{H}}^T}{\bf{Q}}.
 \end{equation}
From (\ref{equ0Rdef43415}) we have ${{\bf{R}}^{l + 1}} = {{\bf{H}}^{l + 1}}{({{\bf{H}}^{l + 1}})^T}{+}k_0^2{{\bf{I}}_{l + 1}}$, into which we substitute (\ref{Haddrow2134}) to obtain
\begin{equation}\label{equ0Riter2nd141}
{{\bf{R}}^{l + 1}} = \left[ {\begin{array}{*{20}{c}}
{{{\bf{R}}^l}}&{{{\bf{p}}_l}}\\
{{\bf{p}}_l^T}&{{{\bf{\bar h}}_{l+1}^T}{\bf{\bar h}}_{l+1} + k_0^2}
\end{array}} \right],
\end{equation}
where ${{\bf{p}}_l}$, a column vector with $l$ entries, satisfies
 \begin{equation}\label{equ0vDefine}
{{\bf{p}}_l} = {{\bf{H}}^l}{\bf{\bar h}}_{l + 1}.
 \end{equation}

The inverse-free recursive algorithm computes ${{\bf{Q}}^{l + 1}}=({{\bf{R}}^{l + 1}})^{-1}$
by equations (11), (16), (13)  and (14) in \cite{Existing_Inverse_free_best_ELM}, which can be written as
 \begin{equation}\label{equ0Qgrow3141}
{{\bf{Q}}^{l + 1}} = \left[ {\begin{array}{*{20}{c}}
{{{{\bf{\tilde Q}}}^l}}&{{{\bf{t}}_l}}\\
{{\bf{t}}_l^T}&{{\tau _l}}
\end{array}} \right],
 \end{equation}
 and
 \begin{subnumcases}{\label{SchuRequ0QplusEntry31}}
{{\bf{\tilde Q}}_l} = {{\bf{Q}}_l} + \frac{{{{\bf{Q}}_l}{{\bf{p}}_l}{\bf{p}}_l^T{{\bf{Q}}_l}}}{{{\bf{\bar h}}_{l + 1}^T{\bf{\bar h}}_{l + 1}^{} + k_0^2 - {\bf{p}}_l^T{{\bf{Q}}_l}{{\bf{p}}_l}}},  &  \label{SchuRequ0QplusEntry31a}\\
{{\bf{t}}_l} =  - \frac{{{{{\bf{\tilde Q}}}_l}{{\bf{p}}_l}}}{{{\bf{\bar h}}_{l + 1}^T{\bf{\bar h}}_{l + 1}^{} + k_0^2}}, & \label{SchuRequ0QplusEntry31b}\\
{\tau _l} = \frac{{{\bf{p}}_l^T{{{\bf{\tilde Q}}}_l}{{\bf{p}}_l}}}{{{{({\bf{\bar h}}_{l + 1}^T{\bf{\bar h}}_{l + 1}^{} + k_0^2)}^2}}} + \frac{1}{{{\bf{\bar h}}_{l + 1}^T{\bf{\bar h}}_{l + 1}^{} + k_0^2}}, & \label{SchuRequ0QplusEntry31c}
\end{subnumcases}
respectively. Notice that in (\ref{equ0Qgrow3141}) and
(\ref{SchuRequ0QplusEntry31}),
%and
%(\ref{equ0QplusEntry31}),
 ${{\bf{t}}_l}$
 %defined in (\ref{equ0QplusEntry31b})
is a column vector with $l$ entries, and ${\tau _l}$
%defined in (\ref{equ0QplusEntry31a})
is a scalar.

\section{Proposed Inverse-Free ELM Algorithms}

Actually the inverse-free recursive algorithm by
(\ref{equ0Qgrow3141}) and
(\ref{SchuRequ0QplusEntry31})
had been mentioned in previous literatures~\cite{Matrix_Handbook,Matrix_SP_book,GlobecommRecursiveVBLAST, TransWC2009ReCursiveBlast,zhfICC2009},
before it was deduced in \cite{Existing_Inverse_free_best_ELM}
by utilizing the Sherman-Morrison formula and the Schur
complement.
That inverse-free recursive algorithm
 can be regarded as the application of the block matrix inverse lemma~\cite[p.30]{Matrix_Handbook},
 and was
called
%represented as
the lemma for inversion of block-partitioned matrix \cite[Ch. 14.12]{Matrix_SP_book}, \cite[equation (16)]{GlobecommRecursiveVBLAST}.
To develop multiple-input multiple-output (MIMO) detectors,
 the inverse-free recursive algorithm
was applied in \cite{GlobecommRecursiveVBLAST, TransWC2009ReCursiveBlast}, and
 its improved version was utilized  in \cite{zhfICC2009, TransComm2010ReCursiveGstbc}.
 \begin{table}[!t]
\renewcommand{\arraystretch}{1.3}
\newcommand{\tabincell}[2]{\begin{tabular}{@{}#1@{}}#2\end{tabular}}
\caption{Comparison of Flops among the Existing and Proposed ELM Algorithms} \label{table_example} \centering
\begin{tabular}{c|c|c|}
                        & {\bfseries   \tabincell{c}{Updating the \\ Intermediate Results}}  & \bfseries  \tabincell{c}{Updating the  \\ Output Weights}      \\
\hline
\bfseries Existing Alg. & $16lK$ & $2MlK$   \\
\hline
\bfseries  Proposed Alg.  $1$   & $6lK$  & $2MK$   \\
\hline
 \bfseries Proposed Alg.  $2$ & $2lK$ & $2MK$  \\
\hline
 \bfseries Proposed Alg.  $3$ & $2lK$ & $2MK$  \\
\hline
\end{tabular}
\end{table}

\subsection{Derivation of Proposed ELM Algorithms}

In the improved version~\cite{zhfICC2009, TransComm2010ReCursiveGstbc},  equation (\ref{SchuRequ0QplusEntry31}) has been simplified into~\cite[equation
(20)]{zhfICC2009}
%can be directly utilized
%which can be written as
\begin{subnumcases}{\label{equ0QplusEntry31}}
{{\tau _l}}  = 1/\left( {({{{\bf{\bar h}}_{l+1}^T}{\bf{\bar h}}_{l+1} + k_0^2})  - {\bf{p}}_l^T
{{\bf{Q}}^{l}}
{\bf{p}}_l } \right) &  \label{equ0QplusEntry31a}\\
{\bf{t}}_l  =  - {{\tau _l}} {{\bf{Q}}^{l}} {\bf{p}}_l & \label{equ0QplusEntry31b}\\
{{{{\bf{\tilde Q}}}^l}}  = {{\bf{Q}}^{l}}  + ({1/{\tau _l}})
{\bf{t}}_l {\bf{t}}_l^T. & \label{equ0QplusEntry31c}
\end{subnumcases}
Accordingly we can utilize (\ref{equ0QplusEntry31}) to simplify (\ref{equ0B_bar94375}) and (\ref{equ0Bwave94375}) into
     \begin{equation}\label{equ1bBarComputeBest93741best}
      {{{\bf{\bar b}}_{l+1}}}={\tau _l}\left(  {\bf{\bar h}}_{l + 1}  -  {{\bf{B}}^{l}} {{\bf{p}}_l} \right)
 \end{equation}
and
 \begin{equation}\label{Bl1computeEffieicient983}
{{\bf{\tilde B}}^l} =  { {{\bf{B}}^{l}}  -{{{\bf{\bar b}}_{l+1}}} {\bf{\bar h}}_{l + 1}^T  {{\bf{B}}^{l}}},
 \end{equation}
 respectively, where ${\tau _l}$ can be computed by
 \begin{equation}\label{taul2hhkhBp321341}
{{\tau _l}}  = 1/\left( {{{{\bf{\bar h}}_{l+1}^T}{\bf{\bar h}}_{l+1} + k_0^2}  -{\bf{\bar h}}_{l + 1}^T
 {{\bf{B}}^{l}}
{{\bf{p}}_l} } \right).
 \end{equation}%[]
Moreover,  from  (\ref{equ1bBarComputeBest93741best})
 and
(\ref{Bl1computeEffieicient983}) we can deduce an  efficient algorithm to update the output weight ${\bf{W}}$, i.e.,
 \begin{equation}\label{W9extendDef232}
 {{\bf{W}}^{l + 1}} = \left[ {\begin{array}{*{20}{c}}
{{{{\bf{\tilde W}}}^l}}&{{\bf{\bar w}}_{l+1}}
\end{array}} \right],
  \end{equation}
  where
  \begin{subnumcases}{\label{WplusSimplest41132MyB}}
{{\bf{\tilde W}}^l} = { {{\bf{W}}^{l}}  - {\bf{\bar w}}_{l + 1} {\bf{\bar h}}_{l + 1}^T  {{\bf{B}}^{l}}} &  \label{WplusSimplest41132aMyB}\\
{\bf{\bar w}}_{l+1} = {\bf{Y}}{{{\bf{\bar b}}_{l+1}}}. & \label{WplusSimplest41132bMyB}
\end{subnumcases}
  The derivation of  (\ref{equ1bBarComputeBest93741best})-(\ref{WplusSimplest41132MyB}) is in Appendix A.

To further reduce the computational complexity,
we can
 update
the unique inverse ${\bf{Q}}$ by
(\ref{equ0vDefine}),
(\ref{equ0QplusEntry31})
and
(\ref{equ0Qgrow3141}),
and  update the output weight ${\bf{W}}$ by  (\ref{W9extendDef232})
where
 \begin{subnumcases}{\label{WplusSimplest41132MyB2myQ}}
{{\bf{\tilde W}}^l} = { {{\bf{W}}^{l}}  + ({\bf{\bar w}}_{l + 1}/{{\tau _l}}) {\bf{t}}_l^T},    &  \label{W2WwpQ321434best}\\
{\bf{\bar w}}_{l+1} = {\tau _l}\left( {\bf{Y}} {\bf{\bar h}}_{l + 1}  - {{\bf{W}}^{l}} {{\bf{p}}_l} \right) & \label{WplusSimplest41132bMyB2myQ}
\end{subnumcases}
are computed from ${\bf{t}}_l$ and ${{\tau _l}}$ in ${{\bf{Q}}^{l + 1}}$.
 The derivation of
     (\ref{WplusSimplest41132MyB2myQ})
 is also in Appendix A.

\begin{table*}[!t]
\renewcommand{\arraystretch}{1.3}
\newcommand{\tabincell}[2]{\begin{tabular}{@{}#1@{}}#2\end{tabular}}
\caption{Experimental Results of the Existing and Proposed Algorithms for Regression Problems} \label{table_example} \centering
\begin{tabular}{|c|c|c c c c|c c c c|c c c c|c|}
\hline
\bfseries Dataset+  & \bfseries Node    &\multicolumn{4}{c|}{{\bfseries Weight Error}}   &\multicolumn{4}{c|}{{\bfseries Output Error (training)}} &\multicolumn{4}{c|}{{\bfseries Output Error (testing)}} & \bfseries Testing \\
\bfseries Kernel  &\bfseries Number & \cite{Existing_Inverse_free_best_ELM} & Alg. 1     & Alg. 2  & Alg. 3  & \cite{Existing_Inverse_free_best_ELM} & Alg. 1     & Alg. 2  & Alg. 3   & \cite{Existing_Inverse_free_best_ELM} & Alg. 1     & Alg. 2  & Alg. 3  & \bfseries MSE  \\
\hline
\hline
\bfseries Airfoil  & 3   & 6e-16  & 8e-16 & 6e-16   & 6e-16   & 8e-15  & 1e-14 & 1e-14   & 1e-14  & 4e-15  & 7e-15   & 5e-15 & 5e-15 & 4.8e-2\\
\bfseries +  & 100   & 2e-11  & 3e-11 & 1e-8   & 2e-11  & 5e-12  & 8e-12 & 4e-9   & 5e-12  & 3e-12  & 5e-12   & 2e-9 & 2e-12 & 1.1e-2\\
\bfseries Gaussian  & 500   & 2e-9  & 6e-10 & 4e-6   & 2e-10   & 1e-10  & 5e-11 & 3e-7   & 2e-11  & 7e-11  & 3e-11   & 2e-7 & 1e-11 & 7.7e-3\\
\hline
\bfseries Energy  & 3   & 2e-14 & 1e-14  & 1e-14  & 7e-15   & 7e-14  & 5e-14 & 4e-14   & 4e-14  & 3e-14  & 2e-14   & 2e-14 & 2e-14 & 3.0e-2\\
\bfseries +  & 100   & 3e-11  & 5e-11 & 4e-8   & 2e-11  & 5e-12  & 6e-12 & 5e-9   & 3e-12  & 3e-12  & 4e-12   & 3e-9 & 1e-12 & 5.0e-3\\
\bfseries Sigmoid  & 500    & 2e-9  & 3e-10 & 1e-6   & 1e-10  & 1e-10  & 2e-11 & 6e-8   & 7e-12  & 6e-11  & 1e-11   & 4e-8 & 4e-12 & 3.7e-3\\
\hline
\bfseries Housing  & 3   & 3e-16  & 4e-16 & 7e-16   & 5e-16   & 2e-15  & 3e-15 & 5e-15   & 4e-15  & 1e-15  & 2e-15   & 3e-15 & 2e-15 & 8.6e-2\\
\bfseries +  & 100   & 2e-12  & 3e-12 & 6e-10   & 1e-12  & 1e-12  & 9e-13 & 3e-10   & 5e-13  & 1e-12  & 3e-12   & 6e-10 & 7e-13 & 7.3e-3\\
\bfseries Sine  & 500   & 4e-10  & 6e-11 & 4e-8   & 2e-11   & 5e-11  & 7e-12 & 6e-9   & 3e-12  & 4e-10  & 7e-11   & 4e-8 & 3e-11 & 5.4e-3\\
\hline
\bfseries Protein  & 3   & 2e-15  & 3e-15 & 8e-16   & 9e-16   & 5e-14  & 6e-14 & 3e-14   & 3e-14  & 2e-14  & 3e-14   & 1e-14 & 1e-14 & 1.8e-1\\
\bfseries +  & 100   & 2e-11  & 2e-11    & 2e-9  & 3e-11  & 4e-11 & 5e-11   & 4e-9  & 6e-11  & 2e-11   & 2e-11 & 2e-9 & 3e-11 & 5.6e-2\\
\bfseries Triangular  & 500   & 2e-9  & 1e-9 & 3e-6   & 1e-9   & 1e-9  & 1e-9 & 2e-6   & 1e-9  & 9e-10  & 7e-10   & 1e-6 & 6e-10 & 4.9e-2\\
\hline
\end{tabular}
\end{table*}

Since the processor units  are limited in  precision,
the  recursive algorithm utilized to update
%the unique inverse
 ${\bf{Q}}$
%of any recursion
may introduce numerical instabilities,
%~\cite{TransSP2003Blast},
 which
occurs only after a very large number of iterations~\cite{TransSP2003Blast}.
Thus instead of the inverse of ${\bf{R}}$ (i.e., ${\bf{Q}}$), we can also
%also develop the proposed algorithm $3$,
%which
update
%the inverse ${\bf{LD}}{{\bf{L}}^T}$
%   factor of ${\bf{R}}$ (
the inverse ${\bf{LD}}{{\bf{L}}^T}$
   factors~\cite{zhfVTC2010DivFree} of ${\bf{R}}$,
   since usually the ${\bf{LD}}{{\bf{L}}^T}$
   factorization is numerically stable~\cite{Matrix_Computations_book}.
   The inverse ${\bf{LD}}{{\bf{L}}^T}$
   factors include the upper-triangular ${{\bf{L}}}$
   and the diagonal ${\bf{D}}$, which
   satisfy
\begin{equation}\label{LDL2QRinv43053}
 {\bf{LD}}{{\bf{L}}^T} = {\bf{Q}} = {{\bf{R}}^{ - 1}}.
  \end{equation}
  From (\ref{LDL2QRinv43053}) we can deduce
 \begin{equation}\label{LDL2QRinv43053Inverse}
 {{\bf{L}}^{-T}}{{\bf{D}}^{-1}}{{\bf{L}}^{-1}}  = {{\bf{R}}},
  \end{equation}
  where the lower-triangular ${{\bf{L}}^{-T}}$ is the conventional ${\bf{LD}}{{\bf{L}}^T}$ factor~\cite{Matrix_Computations_book} of ${{\bf{R}}}$.

 The inverse ${\bf{LD}}{{\bf{L}}^T}$
   factors can be computed from ${\bf{R}}$ directly by
 the inverse ${\bf{LD}}{{\bf{L}}^T}$ factorization in \cite{zhfVTC2010DivFree},
 %which can be written as
 i.e.,
 \begin{subnumcases}{\label{LDLdefineLD133}}
{{\bf{L}}^{l + 1}} = \left[ {\begin{array}{*{20}{c}}
{{{\bf{L}}^l}}&{{{{\bf{\tilde t}}}_l}}\\
{{\bf{0}}_l^T}&1
\end{array}} \right] &  \label{LDLdefineLD133L}\\
{{\bf{D}}^{l + 1}} = \left[ {\begin{array}{*{20}{c}}
{{{\bf{D}}^l}}&{{\bf{0}}_l^{}}\\
{{\bf{0}}_l^T}&{{\tau _l}}
\end{array}} \right],  & \label{LDLdefineLD133D}
\end{subnumcases}
where
\begin{subnumcases}{\label{LDLdefineLD133td}}
{{{\bf{\tilde t}}}_l}= - {{\bf{L}}^l} {{\bf{D}}^l} ({{\bf{L}}^l})^T  {\bf{p}}_l &  \label{LDLdefineLD133t}\\
{{\tau _l}}= 1/\left( {({{{\bf{\bar h}}_{l+1}^T}{\bf{\bar h}}_{l+1} + k_0^2})  - {\bf{p}}_l^T
{{\bf{L}}^l} {{\bf{D}}^l} ({{\bf{L}}^l})^T
{\bf{p}}_l } \right). & \label{LDLdefineLD133VECd}
\end{subnumcases}

We can show that
 ${{{\bf{\tilde t}}}_l}$ in (\ref{LDLdefineLD133t}) and ${\bf{t}}_l$  in (\ref{equ0QplusEntry31b})
  satisfy
\begin{equation}\label{tildet2tTau32943}
{{{\bf{\tilde t}}}_l}={\bf{t}}_l/{{\tau _l}},
 \end{equation}
and ${{\tau _l}}$ in (\ref{LDLdefineLD133VECd}) is equal to ${{\tau _l}}$ in (\ref{equ0QplusEntry31a}),
 by substituting (\ref{LDL2QRinv43053}) into (\ref{LDLdefineLD133t}) and (\ref{LDLdefineLD133VECd}), respectively.
After updating ${\bf{L}}$ and ${\bf{D}}$,
%the inverse ${\bf{LD}}{{\bf{L}}^T}$
%   factor of ${\bf{R}}$ (
we compute the output weight ${\bf{W}}$ by
(\ref{WplusSimplest41132bMyB2myQ}),
%WplusSimplest41132bMyB
 \begin{equation}\label{W2WwpQ321434best4LDL}
{{\bf{\tilde W}}^l} = { {{\bf{W}}^{l}}  + {\bf{\bar w}}_{l + 1} {\bf{\tilde t}}_l^T},
\end{equation}
%(\ref{W2WwpQ321434best4LDL})
and
(\ref{W9extendDef232}),
where   (\ref{W2WwpQ321434best4LDL}) is deduced by
  substituting (\ref{tildet2tTau32943}) into (\ref{W2WwpQ321434best}).

\subsection{Summary and Complexity Analysis of ELM Algorithms}

Firstly let us summarize the existing and proposed inverse-free ELM algorithms,
which all compute the output ${\bf{Z}}$ by
(\ref{equ08}), and compute the estimation error ${\bf{E}}$ by (\ref{Edefine32178a}).
In (\ref{equ08}) and (\ref{Edefine32178a}),  the output weight ${\bf{W}}$ is required.

The existing inverse-free ELM  Algorithm~\cite{Existing_Inverse_free_best_ELM}
uses
(\ref{equ0Bwave94375}),
(\ref{equ0B_bar94375}) and
(\ref{equ0Bextend94375}) to update
the regularized pseudo-inverse
${\bf{B}}$, from which the output weight ${\bf{W}}$ is computed by  (\ref{equ0DefWbyB32797}).
The proposed Algorithm $1$
 uses
 (\ref{equ0vDefine}),
(\ref{taul2hhkhBp321341}),
(\ref{equ1bBarComputeBest93741best}),
(\ref{Bl1computeEffieicient983})
and
(\ref{equ0Bextend94375}) to update
the regularized pseudo-inverse
${\bf{B}}$,
from which
the output weight ${\bf{W}}$ is computed
 by (\ref{WplusSimplest41132MyB})
and
(\ref{W9extendDef232}).
The proposed Algorithm $2$
uses
(\ref{equ0vDefine}),
(\ref{equ0QplusEntry31})
and
(\ref{equ0Qgrow3141}) to
update
the unique inverse ${\bf{Q}}$,
from which
 the output weight ${\bf{W}}$ is computed by
(\ref{WplusSimplest41132MyB2myQ})
%,(\ref{W2WwpQ321434best})
and
(\ref{W9extendDef232}).
The proposed Algorithm $3$
uses
(\ref{equ0vDefine}),
(\ref{LDLdefineLD133td})
and
(\ref{LDLdefineLD133}) to
update
the ${\bf{LD}}{{\bf{L}}^T}$
   factors of ${\bf{Q}}$,
  from which  the output weight ${\bf{W}}$  is computed by (\ref{WplusSimplest41132bMyB2myQ}),
(\ref{W2WwpQ321434best4LDL})
and
(\ref{W9extendDef232}).

\begin{table}[!t]
\renewcommand{\arraystretch}{1.3}
\newcommand{\tabincell}[2]{\begin{tabular}{@{}#1@{}}#2\end{tabular}}
\caption{Speedups in Training Time of the Proposed Algorithms over the Existing Algorithm} \label{table_example} \centering
\begin{tabular}{|c|c|c|c|c|}
\hline
\bfseries Dataset+  & \bfseries Nodes    &\multicolumn{3}{c|}{{\bfseries Speedups}}     \\
\bfseries Kernel  &\bfseries Number  & Alg. 1     & Alg. 2  & Alg. 3    \\
\hline
\hline
\bfseries Airfoil+  & 100   &2.43  &7.99   &5.66  \\
\bfseries Gaussian  & 500   &2.61  &3.96   &2.54  \\
\hline
\bfseries Energy+  & 100   & 2.30 & 4.47  & 3.47  \\
\bfseries Sigmoid  & 500   &  2.51 & 2.32  & 1.55  \\
\hline
\bfseries Housing+  & 100   &2.73  &4.64   &3.32  \\
\bfseries Sine  & 500   &2.77  &1.92   &1.41  \\
\hline
\bfseries Protein+  & 100   &2.54  &19.04   &16.28  \\
\bfseries Triangular  & 500   &2.66  &22.09   &19.29  \\
\hline
\end{tabular}
\end{table}

In the remainder of this subsection, we compare the expected flops (floating-point operations) of the existing ELM algorithm in \cite{Existing_Inverse_free_best_ELM} and
the proposed ELM algorithms.
 Obviously
$l_1 l_3 (2 l_2 - 1) \approx 2 l_1  l_2 l_3$  flops are required
  to multiply a $l_1 \times l_2$ matrix by a $l_2 \times l_3$ matrix,  and  $l_1  l_2$ flops are required to sum two matrices in size $l_1 \times l_2$~\cite{Existing_Inverse_free_best_ELM}.

In Table \Rmnum{1}, we compare the flops of the existing ELM algorithm~\cite{Existing_Inverse_free_best_ELM}
 and the proposed ELM algorithms $1$, $2$ and $3$. As in \cite{Existing_Inverse_free_best_ELM},
 the flops of the existing ELM algorithm do not include the $0(lK)$ entries for simplicity,
 since usually the ELM has large $K$ (the number
of training examples) and $l$ (the number of hidden nodes).
The flops of the proposed ELM algorithms do not include the entries that are $0(lK)$ or $0(MK)$.
Since usually $M/l \approx 0$, it can easily be seen from Table \Rmnum{1} that with respect to the
existing ELM algorithm, the proposed ELM algorithms $1$, $2$ and $3$ only require about
$\frac{3}{8+M}$, $\frac{1}{8+M}$ and $\frac{1}{8+M}$
 of
flops, respectively.

\begin{table*}[!t]
\renewcommand{\arraystretch}{1.3}
\newcommand{\tabincell}[2]{\begin{tabular}{@{}#1@{}}#2\end{tabular}}
\caption{Experimental Results of the Existing and Proposed Algorithms for Classification Problems} \label{table_example} \centering
\begin{tabular}{|c|c|c|c c c c|c c c c|}
\hline
\multirow{2}{*}{\bfseries Dataset }  & \multirow{2}{*}{\bfseries Kernel}   & \multirow{2}{*}{\bfseries Mean/Variance}      &\multicolumn{4}{c|}{{\bfseries Training}}   &\multicolumn{4}{c|}{{\bfseries Testing}}  \\ \cline{4-11}
      &  &   & \bfseries ACC & \bfseries SN    & \bfseries PE  & \bfseries MCC  & \bfseries ACC & \bfseries SN    & \bfseries PE  & \bfseries MCC   \\
\hline
 \multirow{10}{*}{\bfseries MAGIC}  & \multirow{2}{*}{\bfseries Gaussian}   & \bfseries Mean  & 0.8645   & 0.9472   & 0.8584   & 0.6975   & 0.8618   & 0.9459   & 0.8561  &  0.6914   \\
   &     & \bfseries Variance   & 0.0019   & 0.0018   & 0.0018   & 0.0045   & 0.0068   & 0.0058   & 0.0064   & 0.0153  \\
    & \multirow{2}{*}{\bfseries Sigmoid}   & \bfseries Mean  & 0.8602   & 0.9468   & 0.8536   & 0.6877   & 0.8588   & 0.9458   & 0.8525   & 0.6844  \\
   &     & \bfseries Variance   & 0.0019   & 0.0019   & 0.0018   & 0.0044   & 0.0065   & 0.0049   & 0.0063   & 0.0146  \\
    & \multirow{2}{*}{\bfseries Hardlim}   & \bfseries Mean  & 0.8312   & 0.9277   & 0.8315   & 0.6202   & 0.8270   & 0.9249   & 0.8284   & 0.6104  \\
   &     & \bfseries Variance    & 0.0038   & 0.0046   & 0.0045   & 0.0088   & 0.0069   & 0.0066   & 0.0083   & 0.0147  \\
     & \multirow{2}{*}{\bfseries Triangular}   & \bfseries Mean  & 0.8592   & 0.9419   & 0.8555  &  0.6852   & 0.8561   & 0.9398  &  0.8532   & 0.6780  \\
   &     & \bfseries Variance   & 0.0023   & 0.0025   & 0.0024  &  0.0052   & 0.0060   & 0.0051   & 0.0066   & 0.0131 \\
     & \multirow{2}{*}{\bfseries Sine}   & \bfseries Mean   & 0.8640  &  0.9487   & 0.8569  &  0.6966  &  0.8620   & 0.9475   & 0.8552  &  0.6919  \\
   &     & \bfseries Variance   &  0.0017   & 0.0016   & 0.0016  &  0.0040   & 0.0068   & 0.0058   & 0.0061  &  0.0152  \\
\hline
 \multirow{10}{*}{\bfseries Musk}  & \multirow{2}{*}{\bfseries Gaussian}   & \bfseries Mean  &  0.9453   & 0.6791   & 0.9522  &  0.7767   & 0.9412  &  0.6613   & 0.9396  &  0.7586   \\
   &     & \bfseries Variance   &  0.0031   & 0.0193   & 0.0097  &  0.0135   & 0.0070   & 0.0321   & 0.0196   & 0.0238  \\
    & \multirow{2}{*}{\bfseries Sigmoid}   & \bfseries Mean   & 0.9474   & 0.6925  &  0.9539  &  0.7862   & 0.9432   & 0.6745  &  0.9412  &  0.7679   \\
   &     & \bfseries Variance   &  0.0030  &  0.0181   & 0.0097   & 0.0128   & 0.0068   & 0.0308   & 0.0189  &  0.0231  \\
    & \multirow{2}{*}{\bfseries Hardlim}   & \bfseries Mean   & 0.9351   & 0.6185   & 0.9397   & 0.7309   & 0.9299   & 0.5969   & 0.9214  &  0.7075   \\
   &     & \bfseries Variance   &  0.0036   & 0.0216  &  0.0128   & 0.0161   & 0.0076   & 0.0341   & 0.0247   & 0.0268  \\
     & \multirow{2}{*}{\bfseries Triangular}   & \bfseries Mean & 0.9447  &  0.6751   & 0.9528   & 0.7744  &  0.9406  &  0.6579   & 0.9390   & 0.7561   \\
   &     & \bfseries Variance    &  0.0032  &  0.0191  &  0.0099   & 0.0137  &  0.0069  &  0.0318  &  0.0196   & 0.0232  \\
     & \multirow{2}{*}{\bfseries Sine}   & \bfseries Mean  & 0.9462   & 0.6889   & 0.9479    & 0.7808   & 0.9419   & 0.6722   & 0.9326  &  0.7620   \\
   &     & \bfseries Variance   &  0.0025   & 0.0145  &  0.0088   & 0.0105   & 0.0067   & 0.0301  &  0.0173   & 0.0218 \\
\hline
 \multirow{10}{*}{\bfseries Adult}  & \multirow{2}{*}{\bfseries Gaussian}   & \bfseries Mean   &  0.8362  &  0.9321  &  0.8612  &  0.5309  &  0.8359  &  0.9307  &  0.8626 &   0.5259   \\
   &     & \bfseries Variance   &   0.0010   & 0.0018  &  0.0015   & 0.0034  &  0.0012  &  0.0020  &  0.0016  &  0.0041  \\
    & \multirow{2}{*}{\bfseries Sigmoid}   & \bfseries Mean  & 0.8316   & 0.9313  &  0.8569  &  0.5160  &  0.8311   & 0.9297   & 0.8582  &  0.5101   \\
   &     & \bfseries Variance    & 0.0014   & 0.0026   & 0.0023   & 0.0051   & 0.0017   & 0.0027   & 0.0023   & 0.0060  \\
    & \multirow{2}{*}{\bfseries Hardlim}   & \bfseries Mean   &  0.8208  &  0.9314   & 0.8457   & 0.4786  &  0.8200  &  0.9298   & 0.8466  &  0.4711  \\
   &     & \bfseries Variance   &   0.0023   & 0.0038   & 0.0034   & 0.0085  &  0.0026  &  0.0039   & 0.0034   & 0.0094  \\
     & \multirow{2}{*}{\bfseries Triangular}   & \bfseries Mean  & 0.8367   & 0.9338   & 0.8607   & 0.5318   & 0.8366   & 0.9327   & 0.8620  &  0.5270  \\
   &     & \bfseries Variance   & 0.0009   & 0.0018   & 0.0015   & 0.0032   & 0.0012   & 0.0019   & 0.0015   & 0.0040  \\
     & \multirow{2}{*}{\bfseries Sine}   & \bfseries Mean  & 0.8377   & 0.9340   & 0.8616   & 0.5349   & 0.8377   & 0.9330   & 0.8630   & 0.5307   \\
   &     & \bfseries Variance    &  0.0008   & 0.0016   & 0.0014  &  0.0028  &  0.0011   & 0.0017   & 0.0014   & 0.0035  \\
\hline
 \multirow{10}{*}{\bfseries Diabetes}  & \multirow{2}{*}{\bfseries Gaussian}   & \bfseries Mean  &0.7973  &0.6010    &0.7572    &0.5339   &0.7681  &0.5604    &0.7048  &0.4663   \\
   &     & \bfseries Variance   &0.0103  &0.0251    &0.0199    &0.0239    &0.0308  &0.0668    &0.0697  &0.0684  \\
    & \multirow{2}{*}{\bfseries Sigmoid}   & \bfseries Mean &  0.7889  &  0.5746  &  0.7504  &  0.5124  &  0.7738  &  0.5548  &  0.7233  &  0.4781   \\
   &     & \bfseries Variance   & 0.0091   & 0.0209   & 0.0166  &  0.0207  &  0.0312 &   0.0655  &  0.0703  &  0.0693   \\
    & \multirow{2}{*}{\bfseries Hardlim}   & \bfseries Mean  & 0.7673   & 0.5380   & 0.7124   & 0.4602   & 0.7340  &  0.4892  &  0.6515  &  0.3819   \\
   &     & \bfseries Variance   & 0.0159   & 0.0529   & 0.0278   & 0.0402   & 0.0348   & 0.0811   & 0.0775  &  0.0800  \\
     & \multirow{2}{*}{\bfseries Triangular}   & \bfseries Mean  & 0.7964   & 0.5994  &  0.7558   & 0.5317   & 0.7674  &  0.5579  &  0.7046  &  0.4645   \\
   &     & \bfseries Variance   & 0.0103  &  0.0249  &  0.0193   & 0.0238   & 0.0313   & 0.0677   & 0.0709  &  0.0704  \\
     & \multirow{2}{*}{\bfseries Sine}   & \bfseries Mean   & 0.7972   & 0.5912   & 0.7633   & 0.5327   & 0.7721   & 0.5560   & 0.7174  &  0.4742   \\
   &     & \bfseries Variance   & 0.0096   & 0.0228  &  0.0184   & 0.0220   & 0.0306  &  0.0662   & 0.0690  &  0.0679  \\
\hline
\end{tabular}
\end{table*}

Notice that in the proposed ELM algorithm $1$,  ${\bf{\bar h}}_{l + 1}^T
 {{\bf{B}}^{l}}$ computed in (\ref{taul2hhkhBp321341})
 can be utilized in
(\ref{Bl1computeEffieicient983}) and
(\ref{WplusSimplest41132aMyB}).
The dominant computational load of the proposed ELM algorithm $1$ comes from
 (\ref{equ0vDefine}),
(\ref{taul2hhkhBp321341}),
(\ref{equ1bBarComputeBest93741best})
%(\ref{Bl1computeEffieicient983})
and
%(\ref{equ0Bextend94375})
(\ref{WplusSimplest41132bMyB}),
of which the flops are $2Kl$, $2Kl$, $2Kl$ and $2KM$, respectively.
Moreover, in the proposed ELM algorithms $2$ and $3$,
the dominant computational load comes from
 (\ref{equ0vDefine}) and (\ref{WplusSimplest41132bMyB2myQ}),
of which the flops are $2Kl$ and $2KM$, respectively.

 \section{Numerical Experiments}

We follow the simulations in \cite{Existing_Inverse_free_best_ELM},
to compare the existing inverse-free ELM algorithm and the proposed inverse-free ELM algorithms
%are compared
on MATLAB software platform under a Microsoft-Windows Server with  $128$ GB of RAM.
  We utilize a fivefold cross validation to partition the datasets into
training and testing sets. To measure the performance,  we employ  the mean squared error (MSE)
%and the error variance
for  regression problems, and employ  four commonly used indices  for  classification problems, i.e.,
 the prediction accuracy (ACC), the sensitivity (SN), the precision (PE) and the Matthews correlation coefficient (MCC).
 Moreover, the regularization factor is set to $k_0^2=0.1$ to avoid over-fitting.

For the regression problem, we consider energy efficiency dataset~\cite{53_dataSet}, housing dataset~\cite{54_dataSet}, airfoil self-noise dataset~\cite{56_dataSet}, and physicochemical properties of
protein dataset~\cite{51_dataSet}. For those datasets, different activation functions are chosen, which include Gaussian, sigmoid, sine  and
triangular.  As Table \Rmnum{4} in \cite{Existing_Inverse_free_best_ELM}, Table \Rmnum{2} shows the regression performance.
 In table \Rmnum{2}, the weight error and the output error
are defined as $\left\| {\bf{W}}_1-{\bf{W}}_2  \right\|_F$
and $\left\| {\bf{Z}}_1-{\bf{Z}}_2  \right\|_F$, respectively,
where ${\bf{W}}_1$ and ${\bf{Z}}_1$ are computed by an inverse-free ELM algorithm, and
${\bf{W}}_2$ and ${\bf{Z}}_2$ are computed by the
standard ELM algorithm.
We set the initial hidden node number to $2$, and utilize the existing and proposed inverse-free ELM algorithms to
add the hidden nodes one by one till the hidden node number reaches $500$. Table \Rmnum{2} includes
the simulation results for the hidden node numbers $3$, $100$ and $500$.

As observed from Table \Rmnum{2}, after $1$ iteration (i.e., the node number  $3$),
the weight error and the output error are less than $10^{-13}$.
For the existing inverse-free ELM algorithm and the proposed algorithms $1$ and $3$,
the weight error and the output error are less than $10^{-10}$ after $98$ iterations (i.e., the node number  $100$),
and are not greater than $2\times 10^{-9}$ after $498$ iterations (i.e., the node number  $500$). However,
for the proposed algorithms $2$,  the weight error and the output error are not greater than $4 \times 10^{-8}$ after $98$ iterations,
and are not greater than $3 \times 10^{-6}$ after $498$ iterations, since
the  recursive algorithm to update
%the unique inverse
 ${\bf{Q}}$
%of any recursion
 introduces numerical instabilities after a very large number of iterations~\cite{TransSP2003Blast}.
 Overall, the standard ELM, the existing inverse-free ELM algorithm and the proposed ELM algorithms $1$, $2$ and $3$ achieve the same testing MSEs,
 which have been listed in the last column of Table \Rmnum{2}.

\begin{table*}[!t]
\renewcommand{\arraystretch}{1.3}
\newcommand{\tabincell}[2]{\begin{tabular}{@{}#1@{}}#2\end{tabular}}
\caption{Experimental Results of the Existing and Proposed Algorithms on MNIST Dataset} \label{table_example} \centering
\begin{tabular}{|c|c c| c c|c c| c c| c c|}
\hline
  \multirow{2}{*}{\bfseries Algorithm}   & \multicolumn{2}{c|}{{\bfseries Gaussian}}    & \multicolumn{2}{c|}{{\bfseries Sigmoid}}  &  \multicolumn{2}{c|}{{\bfseries Hardlim}} & \multicolumn{2}{c|}{{\bfseries Triangular}}   & \multicolumn{2}{c|}{{\bfseries Sine}}  \\
    & Accuracy   & Speedups  & Accuracy   & Speedups   & Accuracy   & Speedups  & Accuracy   & Speedups   & Accuracy   & Speedups   \\
\hline
  Existing Alg.  & $96.54\%$  &   & $94.67\%$  &   & $93.83\%$  &  & $96.20\%$  &    & $95.85\%$  &   \\
\hline
Proposed  Alg. 1 & $96.54\%$  & 3.41  & $94.67\%$ &  3.77  &  $93.83\%$   &  $3.92$ & $96.20\%$ & $3.14$   & $95.85\%$ & $3.06$  \\
Proposed  Alg. 2 & $96.54\%$  & 45.50  & $94.67\%$ &  44.04  &  $93.83\%$   &  $50.49$ & $96.20\%$ & $43.87$   & $95.85\%$ & $51.73$  \\
Proposed  Alg. 3 & $96.54\%$  & 26.28  & $94.67\%$ &  31.04  &  $93.83\%$   &  $33.69$ & $96.20\%$ & $28.20$   & $95.85\%$ & $34.42$  \\
\hline
\end{tabular}
\end{table*}

The speedups in training time of the proposed ELM algorithms $1$, $2$ and $3$ over the existing inverse-free ELM algorithm are shown in Table \Rmnum{3},
where we add just one node to reach $100$ and $500$ nodes, respectively,
and we do $1000$ simulations to compute the average training time.
 The speedups are computed by
  ${T_{existing}}/{T_{proposed}}$, i.e.,
the  ratio between the training time of the existing ELM algorithm  and that of the proposed ELM algorithm.
  As observed from Table \Rmnum{3}, all the $3$ proposed algorithms
significantly accelerate the existing
inverse-free ELM algorithm.

%We compare the training time for the addition of one node to reach $100$ nodes,
%and that to reach  $500$ nodes.

For the classification problem, we consider MAGIC Gamma telescope
dataset~\cite{58_dataSet}, musk dataset~\cite{57_dataSet}, adult dataset~\cite{60_dataSet} and diabetes dataset~\cite{51_dataSet}. For each dataset, five activation functions are simulated, i.e., Gaussian, sigmoid, Hardlim, triangular and sine.
In the simulations, the standard ELM, the existing inverse-free ELM algorithm and the proposed ELM algorithms $1$, $2$ and $3$ achieve the same performance,
 which have been listed in  Table \Rmnum{4}.

Lastly,  in Table \Rmnum{5} we simulate the existing and proposed algorithms on the
Modified National Institute of Standards and Technology (MNIST)
dataset~\cite{61_dataSet} with $60000$ training images and $10000$ testing images, to show the performance on big data.
To give  the testing accuracy, we set the initial hidden node number to $2000$, and utilize the existing and proposed ELM algorithms to
add hidden nodes one by one till the hidden node number reaches $2200$.   To give the speedups of the proposed algorithms
%$1$, $2$ and $3$
over the existing algorithm,
we compare  the training time to reach $2200$ nodes by
adding one node, and do $500$ simulations to compute the average training time.

As observed from Table \Rmnum{5},  the existing and proposed inverse-free ELM algorithms  bear the same testing accuracy,
while all the $3$ proposed algorithms
significantly accelerate the existing
inverse-free ELM algorithm. Moreover, from Table \Rmnum{5} and Table \Rmnum{3}, it can be seen that usually the proposed algorithm $2$ is faster than the proposed algorithm $3$, and
the proposed algorithm $3$ is faster than the proposed algorithm $1$.

\section{Conclusion}

%is utilized
To reduce the computational complexity of the existing inverse-free ELM algorithm \cite{Existing_Inverse_free_best_ELM},
in this correspondence we utilize the improved recursive algorithm \cite{zhfICC2009, TransComm2010ReCursiveGstbc}
to deduce the proposed  ELM algorithms $1$, $2$ and $3$. The proposed  algorithm $1$ includes a more efficient
inverse-free algorithm to update the regularized pseudo-inverse ${\bf{B}}$. To further reduce the computational complexity,  the proposed  algorithm $2$
 computes  the  output weights directly from the updated inverse ${\bf{Q}}$, and avoids computing the regularized pseudo-inverse ${\bf{B}}$.
Lastly,  instead of updating the inverse ${\bf{Q}}$,
the proposed ELM algorithm $3$ updates the ${\bf{LD}}{{\bf{L}}^T}$
   factors  of the inverse ${\bf{Q}}$ by  the inverse ${\bf{LD}}{{\bf{L}}^T}$ factorization \cite{zhfVTC2010DivFree},
   since the inverse-free recursive algorithm to update the inverse ${\bf{Q}}$
    introduces  numerical instabilities after a very large number of
iterations~\cite{TransSP2003Blast}.
With respect to the
existing ELM algorithm, the proposed ELM algorithms $1$, $2$ and $3$ are expected to require only
%$\frac{2}{4+M}$, $\frac{1}{4+M}$ and $\frac{1}{4+M}$
$\frac{3}{8+M}$, $\frac{1}{8+M}$ and $\frac{1}{8+M}$
 of  flops, respectively. In the numerical experiments, the standard ELM, the
existing inverse-free ELM algorithm and the proposed ELM
algorithms 1, 2 and 3 achieve the same performance in regression and
classification,  while all the $3$ proposed algorithms
significantly accelerate the existing
inverse-free ELM algorithm.  Moreover, in the simulations,  usually the proposed algorithm $2$ is faster than the proposed algorithm $3$, and
the proposed algorithm $3$ is faster than the proposed algorithm $1$.

\appendices
\section{Derivation of  (\ref{equ1bBarComputeBest93741best}),
(\ref{Bl1computeEffieicient983}), (\ref{taul2hhkhBp321341}), (\ref{W9extendDef232}),  (\ref{WplusSimplest41132MyB}) and (\ref{WplusSimplest41132MyB2myQ})}

Substitute
(\ref{Haddrow2134})
and
(\ref{equ0QRrelation3123})
into
(\ref{equ0defB3789})
to obtain
 \begin{equation}\label{equ0defB2HQ984215a}
{{\bf{B}}^{l + 1}} =  \left[ {\begin{array}{*{20}{c}}
({{\bf{H}}^l})^T &  {\bf{\bar h}}_{l + 1}
\end{array}} \right]{{\bf{Q}}^{l + 1}}.
 \end{equation}
Substitute
(\ref{equ0QplusEntry31c})
into
(\ref{equ0Qgrow3141}),
which is then
substituted into (\ref{equ0defB2HQ984215a}) to obtain
 %\begin{equation}\label{equ0defB2HQ984215a2}
${{\bf{B}}^{l + 1}} =  \left[ {\begin{array}{*{20}{c}}
({{\bf{H}}^l})^T &  {\bf{\bar h}}_{l + 1}
\end{array}} \right]\left[ {\begin{array}{*{20}{c}}
{ {{\bf{Q}}^{l}}  + ({1/{\tau _l}})
{\bf{t}}_l {\bf{t}}_l^T }&{{{\bf{t}}_l}}\\
{{\bf{t}}_l^T}&{{\tau _l}}
\end{array}} \right]$,
 %\end{equation}
 i.e.,
 %From  (\ref{equ0defB2HQ984215a2}) we obtain
 \begin{multline}\label{equ0defB2HQ984215a3}
{{\bf{B}}^{l + 1}} =  \\
\left[ {\begin{array}{*{20}{c}}
{{{\left( {({{\bf{H}}^l})^T {{\bf{Q}}^{l}}  + {{\tau _l^{-1}}} ({{\bf{H}}^l})^T
{\bf{t}}_l {\bf{t}}_l^T } + {\bf{\bar h}}_{l + 1} {\bf{t}}_l^T \right)}^T}}\\
{{{\left( ({{\bf{H}}^l})^T
{\bf{t}}_l+  {\tau _l}  {\bf{\bar h}}_{l + 1} \right)}^T}}
\end{array}} \right]^T.
 \end{multline}

To deduce (\ref{equ1bBarComputeBest93741best}),
%(\ref{Bl1computeEffieicient983})
denote the second entry in the right side of
(\ref{equ0defB2HQ984215a3})  as
   \begin{equation}\label{equ1bBarComputeBest93741}
{{{\bf{\bar b}}_{l+1}}}=({{\bf{H}}^l})^T
{\bf{t}}_l+  {\tau _l}  {\bf{\bar h}}_{l + 1},
\end{equation}
into which substitute
(\ref{equ0QplusEntry31b})
to obtain
   \begin{equation}\label{equ1bBarComputeBestbbb2}
 {{{\bf{\bar b}}_{l+1}}}=- {{\tau _l}} ({{\bf{H}}^l})^T
 {{\bf{Q}}^{l}} {{\bf{p}}_l}+  {\tau _l}  {\bf{\bar h}}_{l + 1},
 \end{equation}
 %Finally
 and then substitute
(\ref{equ0defB3789ToQ}) into (\ref{equ1bBarComputeBestbbb2}).

To deduce
%(\ref{equ1bBarComputeBest93741best}),
(\ref{Bl1computeEffieicient983}),
 substitute (\ref{equ0QplusEntry31b}) into
 the first entry in the right side of
 (\ref{equ0defB2HQ984215a3}), and denote it as
${{\bf{\tilde B}}^l}={({{\bf{H}}^l})^T {{\bf{Q}}^{l}}  - ({{\bf{H}}^l})^T
{\bf{t}}_l  {\bf{p}}_l^T  {{\bf{Q}}^{l}} } - {{\tau _l}}{\bf{\bar h}}_{l + 1}  {\bf{p}}_l^T  {{\bf{Q}}^{l}}$,
 i.e.,
    \begin{equation}\label{equ0defB2HQ984215a5}
{{\bf{\tilde B}}^l}={({{\bf{H}}^l})^T {{\bf{Q}}^{l}}  -\left( ({{\bf{H}}^l})^T
{\bf{t}}_l + {\tau _l}  {\bf{\bar h}}_{l + 1} \right) {\bf{p}}_l^T  {{\bf{Q}}^{l}} }.
 \end{equation}
Then substitute (\ref{equ1bBarComputeBest93741}) into (\ref{equ0defB2HQ984215a5})
 %into (\ref{equ0defB2HQ984215a5})
 to obtain
     \begin{equation}\label{equ0defB2HQ984215a6}
{{\bf{\tilde B}}^l}={({{\bf{H}}^l})^T {{\bf{Q}}^{l}}  -{{{\bf{\bar b}}_{l+1}}} {\bf{p}}_l^T  {{\bf{Q}}^{l}} },
 \end{equation}
into which substitute (\ref{equ0vDefine})
%into (\ref{equ0defB2HQ984215a6})
 to obtain
    \begin{equation}\label{equ0defB2HQ984215a7}
{{\bf{\tilde B}}^l}={({{\bf{H}}^l})^T {{\bf{Q}}^{l}}  -{{{\bf{\bar b}}_{l+1}}} {\bf{\bar h}}_{l + 1}^T  ({{\bf{H}}^l})^T  {{\bf{Q}}^{l}} }.
 \end{equation}
 Finally we need to substitute (\ref{equ0defB3789ToQ}) into (\ref{equ0defB2HQ984215a7}).

To deduce (\ref{taul2hhkhBp321341}),
 substitute (\ref{equ0vDefine}) into (\ref{equ0QplusEntry31a}) to obtain
${{\tau _l}}  = 1/\left( {{{{\bf{\bar h}}_{l+1}^T}{\bf{\bar h}}_{l+1} + k_0^2}  -{\bf{\bar h}}_{l + 1}^T ({{\bf{H}}^l})^T
{{\bf{Q}}^{l}}
{{\bf{p}}_l} } \right)$, into which substitute (\ref{equ0defB3789ToQ}).

By
substituting (\ref{Bl1computeEffieicient983}) into
(\ref{equ0Bextend94375}) and
%then
 substituting (\ref{equ0Bextend94375}) into (\ref{equ0DefWbyB32797}),
we can
deduce
(\ref{W9extendDef232}) where ${\bf{\bar w}}_{l+1}$ satisfies (\ref{WplusSimplest41132bMyB}) and
 \begin{equation}\label{WplusSimplest41132aMyBd1}
{{\bf{\tilde W}}^l} = { {\bf{Y}}{{\bf{B}}^{l}}  - {\bf{Y}} {{{\bf{\bar b}}_{l+1}}} {\bf{\bar h}}_{l + 1}^T  {{\bf{B}}^{l}}},
 \end{equation}
 into which
 substitute
(\ref{WplusSimplest41132bMyB})
 and (\ref{equ0DefWbyB32797})
 % into  (\ref{WplusSimplest41132aMyBd1})
 to deduce (\ref{WplusSimplest41132aMyB}).

 To deduce (\ref{W2WwpQ321434best}),
substitute (\ref{equ0defB3789ToQ}) into (\ref{WplusSimplest41132aMyB}) to obtain
${{\bf{\tilde W}}^l} = { {{\bf{W}}^{l}}  - {\bf{\bar w}}_{l + 1} {\bf{\bar h}}_{l + 1}^T  {({{\bf{H}}^l})^T} {{\bf{Q}}^l}  }$, into which substitute
(\ref{equ0vDefine}) to obtain
 \begin{equation}\label{W2WwpQ321434}
{{\bf{\tilde W}}^l} = { {{\bf{W}}^{l}}  - {\bf{\bar w}}_{l + 1} {{\bf{p}}_l^T}   {{\bf{Q}}^l}  },
\end{equation}
and then substitute (\ref{equ0QplusEntry31b}) into (\ref{W2WwpQ321434}).
% to obtain
Moreover, to deduce (\ref{WplusSimplest41132bMyB2myQ}),
 substitute
  % To deduce (\ref{equ1bBarComputeYmultiply394b}),substitute
   (\ref{equ1bBarComputeBest93741best})
    into
    (\ref{WplusSimplest41132bMyB})
    %(\ref{WplusSimplest41132bMyBd1})
  % into (\ref{equ111BarW2Yb329s})
   to obtain
    \begin{equation}\label{equ1bBarComputeYmultiply394}
    {\bf{\bar w}}_{l + 1}={\tau _l}\left( {\bf{Y}} {\bf{\bar h}}_{l + 1}  - {\bf{Y}} {{\bf{B}}^{l}} {{\bf{p}}_l} \right),
 \end{equation}
%Then we can substitute (\ref{equ0DefWbyB32797}) into (\ref{equ1bBarComputeYmultiply394}) to deduce .
into which substitute (\ref{equ0DefWbyB32797}). % to deduce (\ref{WplusSimplest41132bMyB}).

\ifCLASSOPTIONcaptionsoff
  \newpage
\fi

\end{document}